\documentclass{article}

\usepackage{microtype}
\usepackage{subfig}
\usepackage{graphicx}
\usepackage{tikz}
\usepackage{tkz-euclide}
\usepackage{booktabs}
\usepackage{enumitem}

\usepackage{amssymb}
\usepackage{bbold}
\usepackage{mwe}

\usepackage{amsmath}
\usepackage{amsfonts}
\usepackage{amsmath}

\usepackage{hyperref}

\newcommand{\mb}[1]{\mathbf{#1}}
\newcommand{\RNum}[1]{\uppercase\expandafter{\romannumeral #1\relax}}
\definecolor{lightseagreen}{rgb}{0.13, 0.7, 0.67}
\definecolor{dimgray}{rgb}{0.41, 0.41, 0.41}
\definecolor{coralred}{rgb}{1, 0.50, 0.55}
 \definecolor{sienna}{rgb}{0.53, 0.18, 0.09}
 \definecolor{darkslateblue}{rgb}{0.28, 0.24, 0.55} 
\usepackage{multirow}

\newcommand{\STAB}[1]{\begin{tabular}{@{}c@{}}#1\end{tabular}}
\usepackage{icml2020}

\icmltitlerunning{Decoupled Prediction Interval Network}

\begin{document}

\twocolumn[
\icmltitle{Accurate Prediction and Uncertainty Estimation using Decoupled Prediction Interval Networks}



\icmlsetsymbol{equal}{*}

\begin{icmlauthorlist}
\icmlauthor{Kinjal Patel}{1}
\icmlauthor{Steven Waslander}{2}

\end{icmlauthorlist}

\icmlaffiliation{1}{University of Waterloo}
\icmlaffiliation{2}{University of Toronto}
\icmlcorrespondingauthor{Kinjal Patel}{kinjal.patel@uwaterloo.ca}

\icmlkeywords{Uncertainty, Prediction Intervalm, Machine Learning, ICML}

\vskip 0.3in
]



\printAffiliationsAndNotice{} 

\begin{abstract}
We propose a network architecture capable of reliably estimating uncertainty of regression based predictions without sacrificing accuracy. The current state-of-the-art uncertainty algorithms either fall short of achieving prediction accuracy comparable to the mean square error optimization or underestimate the variance of network predictions. We propose a decoupled network architecture that is capable of accomplishing both at the same time. We achieve this by breaking down the learning of prediction and prediction interval (PI) estimations into a two-stage training process. We use a custom loss function for learning a PI range around optimized mean estimation with a desired coverage of a proportion of the target labels within the PI range. We compare the proposed method with current state-of-the-art uncertainty quantification algorithms on synthetic datasets and UCI benchmarks, reducing the error in the predictions by 23 to 34\% while maintaining 95\% Prediction Interval Coverage Probability (PICP) for 7 out of 9 UCI benchmark datasets. We also examine the quality of our predictive uncertainty by evaluating on Active Learning and demonstrating 17 to 36\% error reduction on UCI benchmarks.

\end{abstract}
\section{Introduction}
\label{introduction}
Recent advancements in deep neural networks (NN) \cite{deeplearning} have empowered a wide variety of applications ranging from weather prediction to autonomous vehicles. Accurate predictions of these networks have facilitated their deployment for low-risk applications, but the lack of reliable uncertainty information has been an issue for sensitive applications. The significant impact of a reliable predictive variance has motivated many different approaches to estimate the uncertainty of network predictions.

The Bayesian neural network (BNN) has provided a general framework to represent output prediction with probabilistic distribution instead of point estimates. The prohibitive computational requirements of BNN limits the scalability of the algorithm, thus inspiring successive approximations. While some of these simplified approaches have been able to scale with a complex NN, they often tend to underestimate the uncertainty of the predictions. Alternatively, other algorithms have proposed to model the output prediction as a probability distribution to characterize the variance or model the uncertainty information using PIs. While these methods have been moderately successful in capturing variance information, they often fail to achieve the best possible accuracy for network predictions.
 
\begin{figure}[t]
    \centering
    \subfloat[RMSE vs LL]
    {
        \label{rmse.fig}
        \includegraphics[width=1.55in]{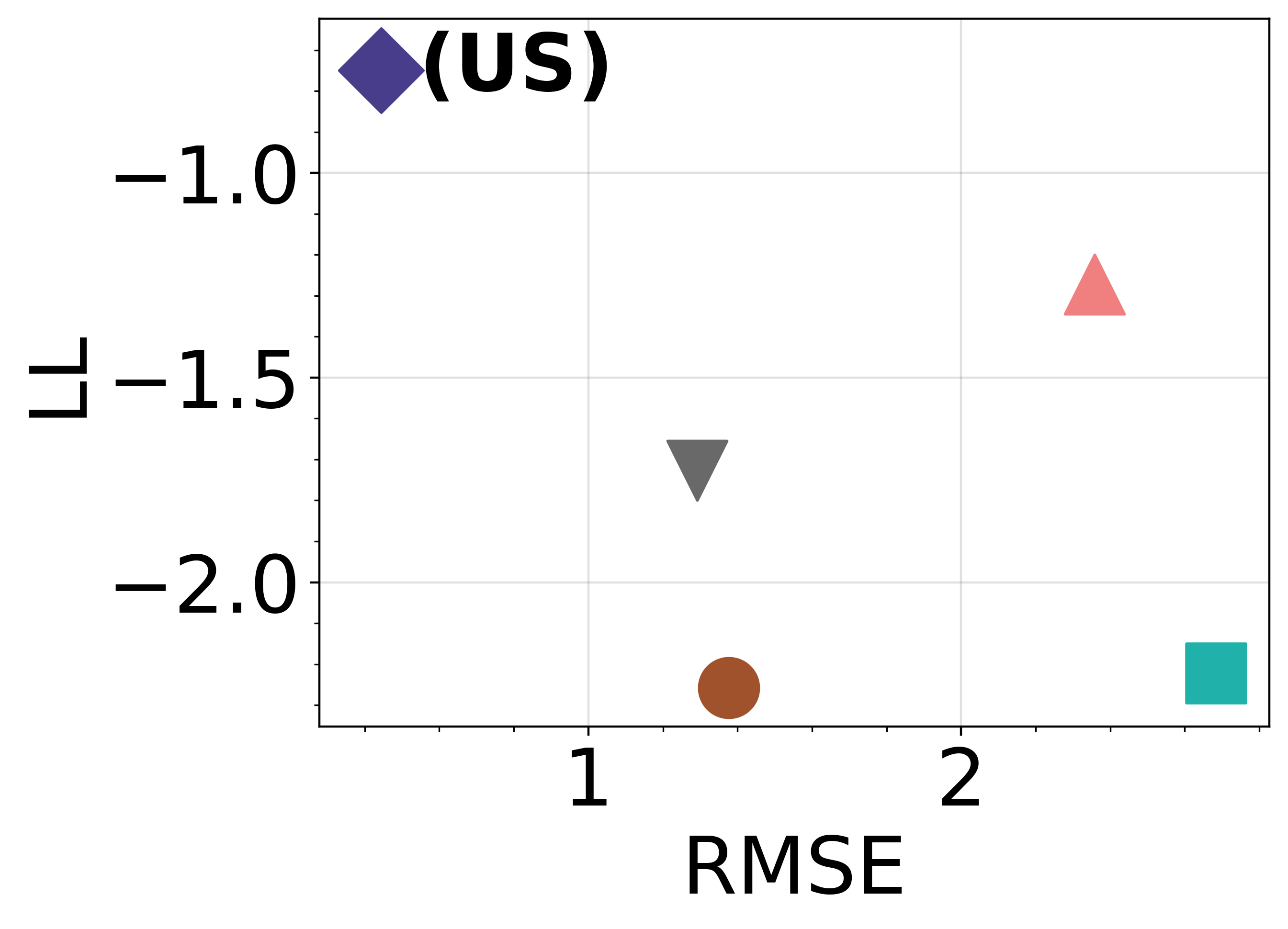}
    }
    \subfloat[PI vs PICP]
    {    
        \label{picp.fig}
        \includegraphics[width=1.53in]{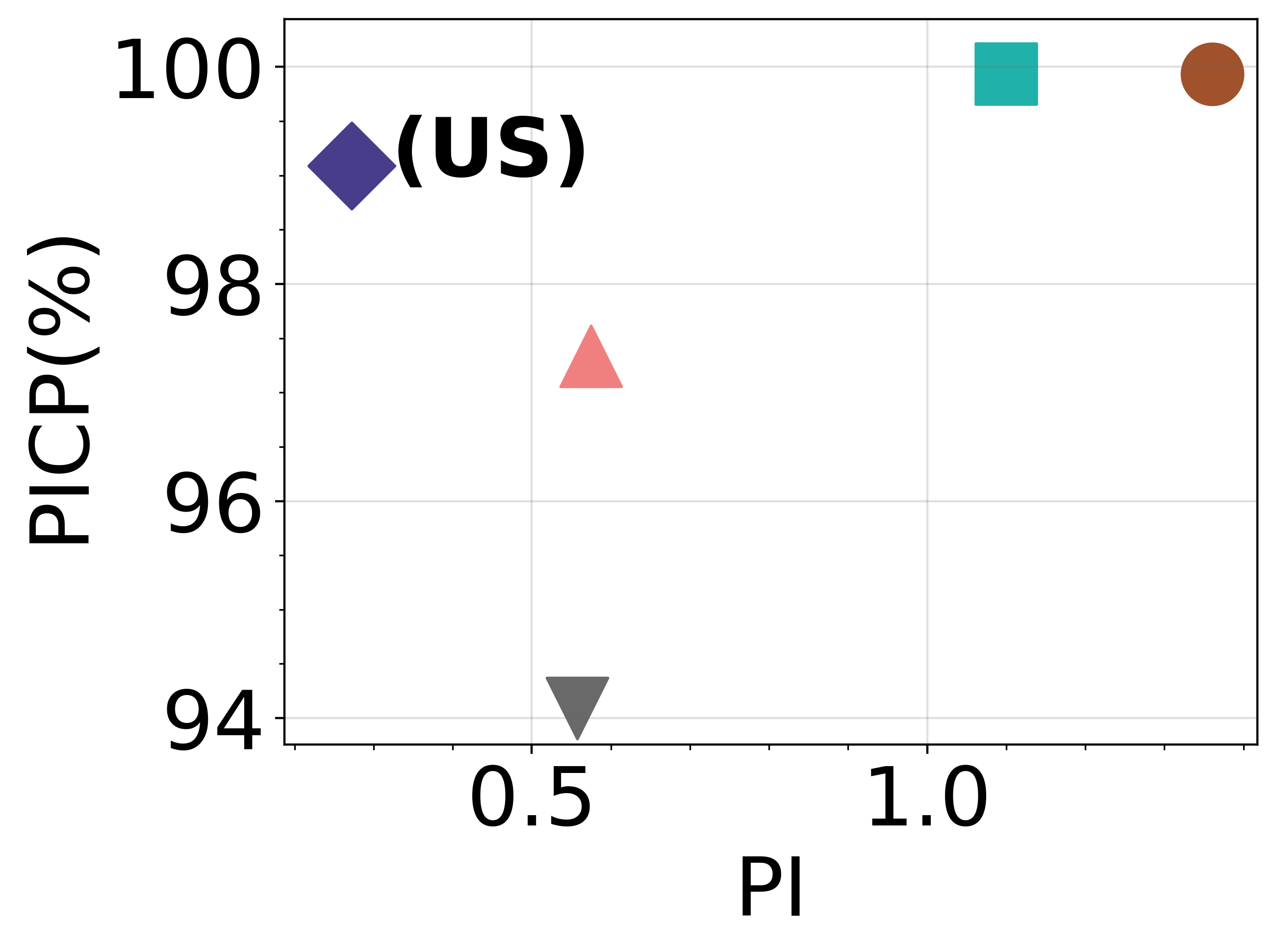}
    }
    {\\
    \vspace{0.3cm}
    \begin{tikzpicture}[scale=0.1]
\fill [coralred] (0,0) -- (2,0) -- (1,2) -- (0,0);

\end{tikzpicture} DeepEns
     \quad
     \begin{tikzpicture}[scale=0.1]
\fill [dimgray] (0,0) -- (-1,2) -- (1,2) -- (0,0);

\end{tikzpicture} MC-Dr
     \quad
     \begin{tikzpicture}[scale=0.1]
\fill [lightseagreen] (0,0) -- (2,0) -- (2,2) -- (0,2) -- (0,0);

\end{tikzpicture} Quality-PI  \quad   \begin{tikzpicture}[scale=0.1]
\fill [sienna](2,2) circle (1cm);

\end{tikzpicture} Split-Train \quad
     \begin{tikzpicture}[scale=0.1]
\fill [darkslateblue] (0,0) -- (-1,1) -- (0,2) -- (1,1) -- (0,0);

\end{tikzpicture} DPIN}
    \caption{Comparison of RMSE vs LL and PI vs PICP(\%) for \texttt{Energy} dataset from UCI benchmark. \subref{rmse.fig} With the least RMSE and the highest log likelihood, left most corner would be the best possible estimate. \subref{picp.fig} A point to the left most side above 95\% mark would be the most favorable position capturing the least PI with $PICP \geq 95\%$.}
    \label{fig:intro_rmse}
    \vspace{-0.2in}
\end{figure}
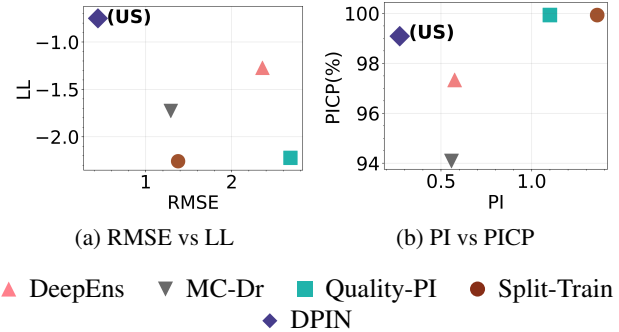
We capture this dichotomy in Figure \ref{fig:intro_rmse} for five different algorithms, DeepEns \cite{deepensemble}, MC-Dr \cite{mcdropout}, Quality-PI \cite{highquality}, Split-Train \cite{reliable} and DPIN (us) on the \texttt{Energy} dataset from UCI benchmark \cite{uci}. We evaluate the performance of these algorithms with four metrics:  root mean square error (RMSE) vs log-likelihood (LL) (Figure 1a), and prediction interval coverage percentage (PICP) vs PI (Figure 1b). Note that the value of RMSE is aimed to be as small as possible, while the LL estimations, assuming Gaussian distribution, is expected to be as high as possible. PIs encapsulating an upper and a lower bound for a particular prediction is expected to ensure the target label would lie in the predicted range with a high probability (95\% or 99\%) representing PICP. The desired value of PICP in this work is 95\% while maintaining the PI to be as small as possible.

As demonstrated by the RMSE-LL comparison (Figure \ref{rmse.fig}), the prediction accuracy attained by Split-Train and DeepEns network is often bottle-necked owing to limitations of the negative log-likelihood (NLL) loss optimization \cite{decouple_nix}. Furthermore, the distribution assumption about data noise would overestimate the PI of the network predictions with asymmetric noise distribution \cite{highquality}. While the Quality-PI algorithm has showcased an ability to eliminate distribution assumption of data noise, it still suffers from reduced prediction accuracy owing to the selection of mid-point of the PI for target prediction instead of explicitly learning the ground truth. 

Quantitatively, The RMSE vs LL comparison (Figure \ref{rmse.fig}) demonstrate a noticeable RMSE reduction of 68\% - 84\% achieved by our proposed method (DPIN) while improving the LL by 41\% - 67\% compared to the current state-of-the-art methods for the \texttt{Energy} dataset. The improvement of uncertainty estimations showcased by higher LL is further validated by attaining 95\% PICP with reduced PI of 53\% - 80\% compared to other algorithms on the \texttt{Energy} dataset and presented via the PI-PICP comparison in the Figure \ref{picp.fig}.

The proposed DPIN method is able to outperform the state-of-the-art methods by employing the mean square error (MSE) loss for optimizing the network predictions and adding a decoupled network for estimating optimal PI range. We derive a custom loss function for learning the PI range, thus eliminating the distribution assumption about data noise. Similar to Quality-PI, we extend this custom loss function to encapsulate target predictions in the estimated PI range with a probability stated by PICP. We optimize the network using a two-stage learning process. Stage I includes learning of optimal network predictions using MSE Loss minimization and stage II models optimal PI range using derived custom loss function without altering network predictions. 

We summarize the main contributions of our work below :
\begin{enumerate}[noitemsep]
\vspace{-0.1in}
\item An upto-date study of uncertainty estimation methods and current standing of the state-of-the-art.
\item A novel decoupled training approach that breaks away from the limiting simultaneous optimization of target predictions and uncertainty estimations.
\item A custom distribution free loss function for estimating PI with respect to explicit target prediction.
\item Empirical evaluation of our proposed method on a variety of tasks, achieving an improvement of 17-36\% for active learning tasks over state-of-the-art algorithms, averaged over all datasets. Furthermore, we achieve 23-34\% improvement in RMSE for regression related tasks while maintaining 95\% PICP on 7 out of 9 datasets.
\end{enumerate}

\section{Related Work}
We now present a study of existing uncertainty estimation algorithms and their contribution while noting their limitations to build motivation for DPIN.

\textbf{Bayesian Neural Network}\\
The merits of encapsulating the NN output in a probabilistic distribution have encouraged researchers to investigate many different approaches. Earlier efforts on Bayesian methods \cite{bnn_mackay, neal1995} revolved around defining a prior distribution over network parameters and learning the posterior distribution which assisted in estimating the predictive uncertainty. The non-trivial task of choosing an informative prior and high computational cost of estimating the posterior distribution in a deep network ensures that exact Bayesian methods remain intractable for practical purposes. In order to tackle these limitations, a variety of approximations have emerged.

\textbf{Practical Variational Inference}\\
Practical variational inference \cite{Graves2011} optimizes the variational lower bound by employing Monte-Carlo approximation. It enables the algorithm to scale beyond single layer NNs, but the noise from stochastic computation with Monte-Carlo approximation leads to poor performance in practical applications. The reparameterization trick \cite{kingma_autoencode} mitigates those limitations and achieves better performance on a moderately sized network. The parametric Bayesian inference \cite{weight, PBP} places independent prior distribution over network parameters and optimizes the posterior parameter distribution to sample diverse predictions using parameter variances. A Gaussian distribution over the parameters would at least double the number of parameters required by the parametric Bayesian network compared to a standard NN, which becomes computationally expensive for complex deep NNs \cite{bayescnn}.

\textbf{Scalable Variational Inference}\\
Another practical approximation of variational inference is realized by employing Monte Carlo Dropout (MC-Dr) \cite{mcdropout} to generate ensembles of the network by performing multiple forward passes with dropouts at inference time. The ubiquity and simplicity of adapting to MC-Dr has made it a popular choice for estimating predictive uncertainty compared to other dropout based approximations \cite{VIdr, bayesdr}. The MC-Dr is further extended \cite{mcdr_kendall} to combine epistemic uncertainty with aleatoric uncertainty generating a complete solution to estimate predictive uncertainty. In more recent work, the popularity of batch normalization has inspired researchers to investigate the applicability of Monte Carlo Batch Normalization (MCBN) to approximate inference in Bayesian models \cite{mcbn}. However, these methods often tend to underestimate the predictive uncertainty in many cases, particularly out of distribution examples.

\textbf{Ensemble Methods}\\
An alternative solution, DeepEns \cite{deepensemble}, proposes to train multiple networks with different parameter initialization to generate a diverse ensemble of the NNs. The combination of MVE \cite{MVE} estimating aleatoric uncertainty and the ensemble of the NNs capturing epistemic uncertainty is able to encapsulate the predictive variance on test examples with known distribution as well as out-of-distribution. Instead of training multiple networks, HydraNet \cite{hydranet} devises a network architecture comprising a base network connecting to various smaller networks. It has showcased that these multiple network heads with different initialization can produce diverse predictions estimating reliable predictive uncertainty. However, both of these approaches have demonstrated poor performance in terms of accuracy of their predictions, when compared to other state-of-the-art methods owing to their reliance on NLL optimization. 

Another approach called anchored ensembling \cite{bayesian_ensemble} has proposed a randomized MAP sampling (RMS) procedure to realize approximate Bayesian ensembling. A distribution of MAP solution approximating true posterior can be obtained by injecting noise to either target labels or regularization of the network parameters. Due to a lack of reliable noise injection procedure for complex deep networks, anchored ensembling regularizes the network parameters around prior distribution instead of anchor distribution (which presents maximum likelihood covariance of the parameters) to realize practical Bayesian ensembling. However, the quality of predictive uncertainty obtained using the custom RMS procedure depends on the non-trivial tasks of selecting a reliable prior distribution.

\textbf{Split Training}\\
In order to avoid ensemble methods, Split-Train \cite{reliable} has employed a mean and variance network \cite{MVE} along with modified mini-batch training. It models the variance of the prediction using a gamma distribution to attain a similar variance extrapolation to Gaussian processes. It takes advantage of the euclidean distance between a test sample and the training data to determine the familiarity of the network with the input at hand. It would assist in estimating the confidence of the network in the prediction, by producing lower variance for input closer to training data and high variance for out-of-distribution input samples. The computation of test input similarity with training data is based on the euclidean distance which can be difficult to estimate for high dimensional input features.

\textbf{High Quality PI}\\
Instead of assuming any output distribution, the optimization of quality metrics (e.g., PICP, mean prediction interval width (MPIW)) has enabled NNs to produce accurate PIs. A high-quality PI should be as narrow as possible while capturing a desired proportion of the target labels. Lower Upper Bound Estimation(LUBE) \cite{LUBE} has incorporated this principle into a loss function for the first time. However, its incompatibility with gradient descent optimization has encouraged further modifications. The Quality-PI \cite{highquality} derives a custom objective function, Quality-driven distribution-free (QD) Loss, in order to address the limitations of LUBE and construct an algorithm which can be optimized by standard NN optimization. It has demonstrated the ability to learn accurate PIs while maintaining the desired PICP. However, Quality-PI does not explicitly model the target prediction while estimating the PI range. The selection of mid-point of the PI as output prediction attains lower prediction accuracy compared to the state-of-the-art method with L2 loss minimization.

We note that the above discussed methods have not been able to provide a scalable solution that is able to output astute target prediction with reliable predictive uncertainty estimates. With DPIN, we solve this by creating two separate Deep NNs with decoupled learning of target label and PI. We combine this decoupled training with a custom loss function which encapsulates a desired proportion of the target labels within PI bounds that are learnt from the difference between predictions and target labels. Combined, we achieve 17-36\% improvement in prediction accuracy over a range of active learning and regression tasks.
\section{Decoupled Prediction Interval Network (DPIN)}
The predictive uncertainty in deep networks comprises of 1. Epistemic uncertainty and 2. Aleatoric uncertainty. The epistemic uncertainty, often referred to as model uncertainty, indicates the confidence in the current prediction based on known training samples and parameters of the network. MC Dropout\cite{mcdropout} and Deep ensembles\cite{deepensemble} are current state-of-the-art scalable methods that learn the epistemic uncertainty of the model. The aleatoric uncertainty, also known as data uncertainty, represents the irreducible noise resulting from the stochasticity of the data. While MVE \cite{MVE} as well as Quality-PI \cite{highquality} are able to capture the aleatoric uncertainty, they often tend to reduce target prediction accuracy compared to methods optimizing mean square error (MSE). 

In order to achieve state-of-the-art prediction accuracy as well as reliable uncertainty estimation, we propose a Decoupled Prediction Interval Network (DPIN) which can split up the training of predictions and their PI bounds. The advantages of DPIN are twofold: 1. Lockstep improvement in accuracy of the predictions as well as uncertainty estimates, without one affecting the other, and 2. Direct integration with any pre-trained network without altering the trained parameters of the prediction network, enabled by DPIN's decoupled training approach.

\textbf{Preliminaries:}
Let the dataset $\mathcal{D} = \{\mb{x}_i, y_i\}^N_{i=1}$ comprise of i.i.d. observations. The underlying non-linear function of NN is denoted by $\mu$. Let the output prediction of the network be defined by $y_i = \mu{(\mb{x}_i)} + \epsilon_i$, where the additive noise $\epsilon_i$ consists of data noise as well as model noise.  Note that although we are denoting output variable as one-dimensional scalar value, it can easily be extended to multivariate problems. 

\subsection{Aleatoric Uncertainty vs Prediction Accuracy}
\subsubsection{NLL Loss Optimization}
A dedicated NN's output, predicting aleatoric uncertainty (originally proposed by MVE \cite{MVE}) utilizes negative log-likelihood (NLL) function to encapsulate the prediction distribution defined by eq. (\ref{eq:nll}).
\begin{equation}
    \mathcal{L}_\mathrm{NLL} = \frac{1}{N}\sum_{i=1}^{N}\frac{\log{\sigma^2_a(\mb{x}_i)}}{2}+\frac{(y_i-\mu(\mb{x}_i))^2}{2\sigma_a(\mb{x}_i)^2} + constant
    \label{eq:nll}
\end{equation}
As observed in \cite{ensemble, reliable, hydranet}, simultaneous optimization of mean and variance using NLL loss does not always result in comparable accuracy to MSE optimization.
The relatively poor accuracy compared to other state-of-the-art methods is attributed to NLL loss optimization instead of MSE optimization,  as evidenced in \cite{decouple_nix}. The loss function in eq. (\ref{eq:nll}) requires simultaneous optimization of the target prediction ($\mu$) as well as the data variance ($\sigma^2_a$). In order to understand the cause of reduced accuracy, we analyze the Gradient of NLL function for an individual sample $\mb{x}$ w.r.t. mean $\mu(\mb{x})$ and variance $\sigma_a(\mb{x})^2$ in eq. (\ref{eq:grad_nll_mu}) and (\ref{eq:grad_nll_var}) respectively.
\begin{align}
      \frac{\partial\mathcal{L}_\mathrm{NLL}}{\partial \mu(\mb{x})} &= \frac{[\mu(\mb{x})-y]}{\sigma_a(\mb{x})^2}\label{eq:grad_nll_mu}\\
     \frac{\partial\mathcal{L}_\mathrm{NLL}}{\partial \sigma_a(\mb{x})^2} &= \frac{1}{\sigma_a(\mb{x})^4}([\mu(\mb{x})-y]^2 - \sigma_a(\mb{x})^2))
    \label{eq:grad_nll_var}
\end{align}
The gradients replicate the effects of weighted regression, where the weights given to a sample would vary with input features. Effectively, to minimize $\frac{\partial\mathcal{L}_\mathrm{NLL}}{\partial \sigma_a(\mb{x})^2}$ in eq. (\ref{eq:grad_nll_var}), the optimizer would estimate high variance for predictions with larger error and low variance for predictions with a smaller error. It encourages the network to accentuate the learning of low variance predictions while underemphasizing the learning of the predictions with high uncertainty. Early training influence would significantly impact the output prediction accuracy and corresponding noise regions leading to a sub-optimal solution \cite{decouple_nix}.

\subsubsection{QD Loss Optimization}
The QD loss function, proposed by Quality-PI, emphasizes learning the minimum PI while encapsulating the desired proportion of the target labels to represent the aleatoric uncertainty of the data. The optimization of PI is particularly beneficial with asymmetric noise distribution. Since the network is only predicting the upper and lower bound of the PI, the target prediction is assumed to be a mid-point of the PI, which might not be optimal, particularly with asymmetric noise distribution. We confirm that this sub-optimal network prediction indeed leads to high RMSE values for target prediction over the UCI benchmark in our evaluations in section \ref{sec:exper}.

\subsection{Decoupled Network Training}
In order to model reliable aleatoric uncertainty with accurate network predictions, we propose a decoupled PI network, thus splitting the learning of target labels and a distribution-free PI range encapsulating the prediction uncertainty.

\begin{figure}
    \centering
    \subfloat[]
    {
        \label{dpin.fig}
        \includegraphics[width=1.8in]{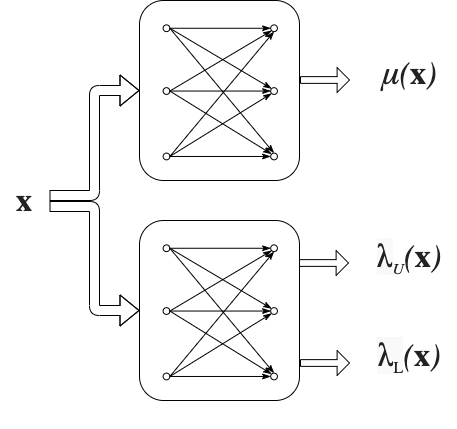}
    }
    \subfloat[]
    {    
        \label{pi.fig}
        \includegraphics[width=1.3in]{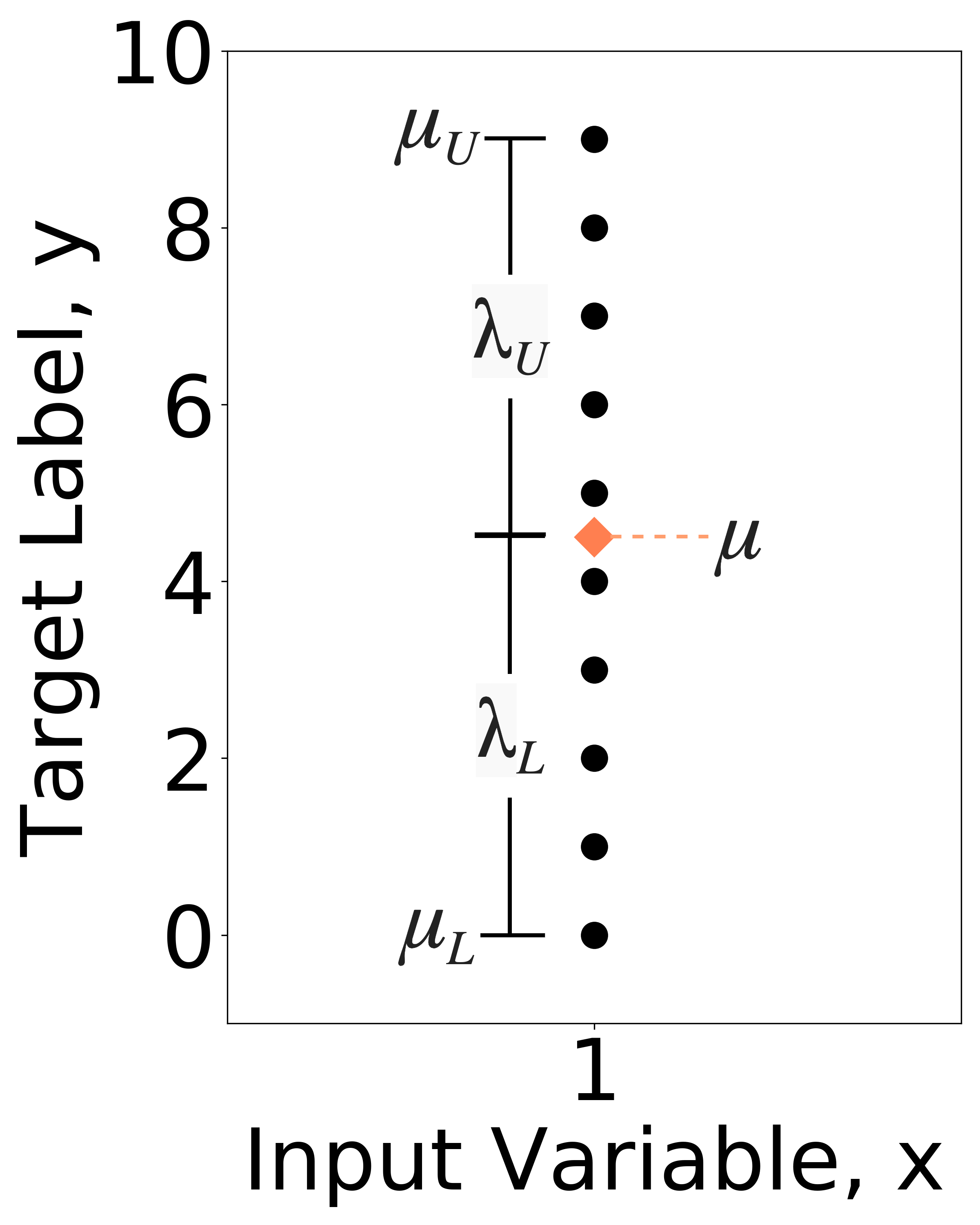}
    }
    \caption{Decoupled Prediction Interval Network: \subref{dpin.fig} NN with outputs representing mean and PI. \subref{pi.fig} Toy example demonstrating optimal prediction $\mu$ with upper bound $\mu_U = \mu + \lambda_U$ and lower bound $\mu_L = \mu - \lambda_L$}
    \label{fig:nn}
    \vspace{-0.2in}
\end{figure}

The proposed network architecture, inspired from \cite{decouple_nix}, with two separate NNs estimating the mean and PI of the prediction is illustrated in Figure \ref{dpin.fig}.  The training of the network is divided into two stages: In stage \RNum{1}, the common base (if it exists) and the network head modeling non-linear function $\mu$ is optimized using MSE loss function described by eq. (\ref{eq:mse_mean}).
\begin{equation}
    \mathcal{L}_{\mu} = \frac{1}{N}\sum_{i=1}^{N}(y_i-\mu(\mb{x}_i))^2
    \label{eq:mse_mean}
\end{equation}
MSE loss function optimization only affects the parameters modeling mean estimates and the common base network (if present), whereas the network head parameters modelling PI estimation are not updated during stage \RNum{1}. Once the target predictions with stage \RNum{1} are optimized, the head network parameters learning PI estimates are optimized to model the upper bound $\mu_U$ and lower bound $\mu_L$ of the PI.

Figure \ref{pi.fig} demonstrates a toy example with multiple target values of $y$ for a single input value of $x$. It also showcases the optimal prediction of $\mu$ with PI range $\lambda_U + \lambda_L$. In this particular case, the optimal width of the upper bound would be $\lambda_U = \max(y) - \mu$ and lower bound would be $\lambda_L = \mu - \min(y)$. While it would be trivial to determine the correspondence of multiple target values for a particular input feature in a single dimension model, the task of finding a correlation between a high dimension input features with multiple target values becomes computationally prohibitive.

We propose to optimize the PI bounds using the difference between network predictions $\mu_i$ and target labels $y_i$ given by eq. (\ref{eq:loss_pi}).
\begin{equation}
    \mathcal{L}_\mathrm{PI} = 
     \begin{cases}
      \sum_{j=1}^{N_1}(\lambda_{U_j} - [y_j - \mu(x_j)])^2 \quad \quad \quad y_j > \mu(x_j) \\[8pt]
      \sum_{j=1}^{N-N_1}((\lambda_{L_j} - [\mu(x_j) - y_j])^2 \quad \quad y_j \leq \mu(x_j)\\
     \end{cases}
    \label{eq:loss_pi}
\end{equation}  
Instead of modelling the upper and lower bound widths of the PI range, $\mathcal{L}_\mathrm{PI}$ learns the upper bound as an average of the target labels, which are greater than the predicted values and lower bound as an average of the ground-truth, which are lesser than the network predictions. This function alone does not capture the desired PI range capturing a high proportion of the target values. In order to encapsulate the required proportion of the target labels within the predicted interval range, the loss function needs to be extended using additional PICP constraints.

The percentage coverage can be modeled using prediction interval(PI), which covers a desired proportion of the observations. The PI can be represented by eq. (\ref{eq:PI}), where $\alpha$ is the desired percentage coverage for observations, common choices for $\alpha$ include $0.95$ or $0.99$.
\begin{equation}
    \mathrm{Pr}(\mu_L \leq y \leq \mu_U) \geq \alpha
    \label{eq:PI}
\end{equation}
The percentage coverage of the prediction interval is described by eq. (\ref{eq:picp}).
\begin{equation}
    \mathrm{PICP} = \frac{1}{N} \sum \mathbb{1}_{y \leq \mu_H} \cdot \mathbb{1}_{y \geq \mu_L}
    \label{eq:picp}
\end{equation}
where, $\mathbb{1}_{y \leq \mu_H}$ and $\mathbb{1}_{y \geq \mu_L}$ are the binary vectors denoting the success or failure of fulfilling the conditions $y \leq \mu_H$ and $y \geq \mu_L$ respectively.

In order to optimize the PI range with the explicit percentage coverage of the target labels within PI estimates, the loss function described by eq. (\ref{eq:loss_picp}) can be optimized.
\begin{equation}
    \mathcal{L}_{\mathrm{PICP}} = (0.95-\mathrm{PICP})^2
    \label{eq:loss_picp}
\end{equation}
While PICP loss $\mathcal{L}_\mathrm{PICP}$ enforces a certain percentage of the target labels within PI bounds, it can overestimate the PI width without any additional constraints. The weighted combination of eq. (\ref{eq:loss_pi}) and (\ref{eq:loss_picp}) encapsulate a complete solution for modelling optimal PI range as defined by eq. (\ref{eq:dpi}), where $\eta_1, \eta_2$ represents weights assigned to the loss function $\mathcal{L}_{\mathrm{PI}}$ and $\mathcal{L}_{\mathrm{PICP}}$.
\begin{equation}
    \mathcal{L}_{\mathrm{DPI}} = \eta_1 \mathcal{L}_{\mathrm{PI}} + \eta_2 \mathcal{L}_{\mathrm{PICP}}
    \label{eq:dpi}
\end{equation}
This two stage training approach allows us to decouple the output estimations and aleatoric uncertainty characterization which can provide improved performance, and easier integration to existing pre-trained models without needing to re-train the complete model. 

\subsection{Epistemic Uncertainty}

To estimate the predictive uncertainty, the epistemic uncertainty of the network, along with the prediction noise, is required to be estimated. A simple and scalable method to obtain the parameter uncertainty can be realized by generating an ensemble of the network either by employing MC Dropout\cite{mcdropout} or training multiple NNs with different initialized parameters. While MC Dropout generates different predictions by enabling dropout at the inference time using a single trained network, it often tends to underestimate the predictive uncertainty, especially with out-of-distribution samples \cite{hydranet, reliable}. The diverse set of networks resulting from different parameter initialization in deep ensembles \cite{deepensemble} has demonstrated improved epistemic uncertainty estimations compared to MC-Dr.

Given an ensemble of M decoupled PI networks, optimal network prediction $\Bar{\mu}$ is obtained via mean of ensemble predictions $\mu_j$ described by eq. (\ref{eq:ens_mu}).
\begin{equation}
\Bar{\mu}(\mb{x}_i) = \frac{1}{M} \sum_{j=1}^{M} \mu_j(\mb{x}_i)
    \label{eq:ens_mu}
\end{equation}
The diversity of the ensemble of the networks can be encapsulated by utilizing the variance of the upper and lower bound of the ensemble PI. The variance of the ensemble PI bound can be obtained using eq. (\ref{eq:var_PI}), where $\Bar{\mu}_{U_j}(x_i) \text{ and } \Bar{\mu}_{L_j}(x_i)$ represent the mean of the ensemble PI bounds ${\mu}_{U_j} \text{ and }, {\mu}_{L_j}$.\vspace{-0.1in}
\begin{equation}
    \begin{split} \sigma^2_{U_j}(\mb{x}_i) &= \frac{1}{M-1} \sum_{j=1}^{M}(\mu_{U_j}(\mb{x}_i) - \Bar{\mu}_{U_j}(\mb{x}_i))^2 \\ \sigma^2_{L_j}(\mb{x}_i) &= \frac{1}{M-1} \sum_{j=1}^{M}(\mu_{L_j}(\mb{x}_i) - \Bar{\mu}_{L_j}(\mb{x}_i))^2
    \end{split}
    \label{eq:var_PI}
\end{equation}
Similar to Quality-PI, we also combine the aleatoric uncertainty and epistemic uncertainty by obtaining upper and lower bound of the ensemble PI using eq. (\ref{eq:ens_PI}).
\begin{equation}
    \begin{split}
     \Tilde{\mu}_U(\mb{x}_i) = \Bar{\mu}_{U_j}(\mb{x}_i) + 1.96 \sigma_{U_j}(\mb{x}_i)\\
     \Tilde{\mu}_L(\mb{x}_i) = \Bar{\mu}_{L_j}(\mb{x}_i) - 1.96 \sigma_{L_j}(\mb{x}_i)
    \end{split}
    \label{eq:ens_PI}
\end{equation}
\section{Experiments}
\label{sec:exper}

This section demonstrates the evaluation of DeepEns \cite{deepensemble}, MC-Dr \cite{mcdropout}, Quality-PI \cite{highquality}, Split-train \cite{reliable}, and the proposed DPIN methods 
by comparing the accuracy and quality of uncertainty estimates. These methods are assessed on a synthetic dataset, and 9 UCI benchmark datasets \cite{uci} for regression and active learning related tasks. Details of network architectures, hyperparameters, and code implementation  can be found in the supplementary material.

\subsection{Synthetic Dataset Regression}
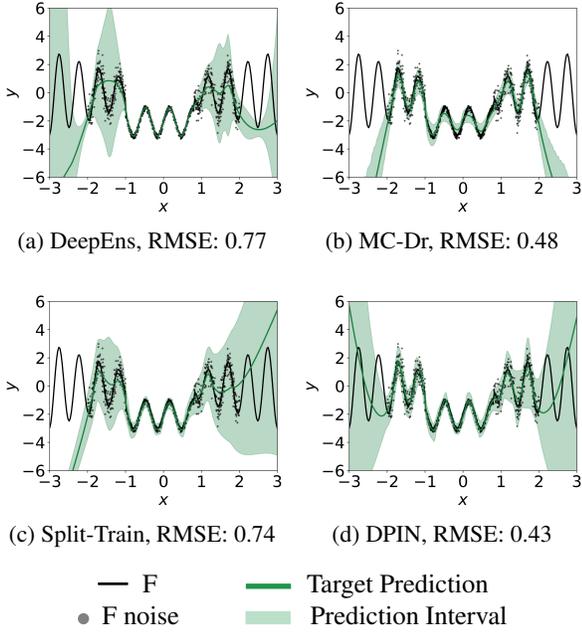
\begin{figure}
    \centering
    \subfloat[DeepEns, RMSE:  0.77]
    {    
        \label{deepens.fig}
        \includegraphics[width=1.5in,height=1.5in,keepaspectratio,trim={0 0 0 2cm},clip]{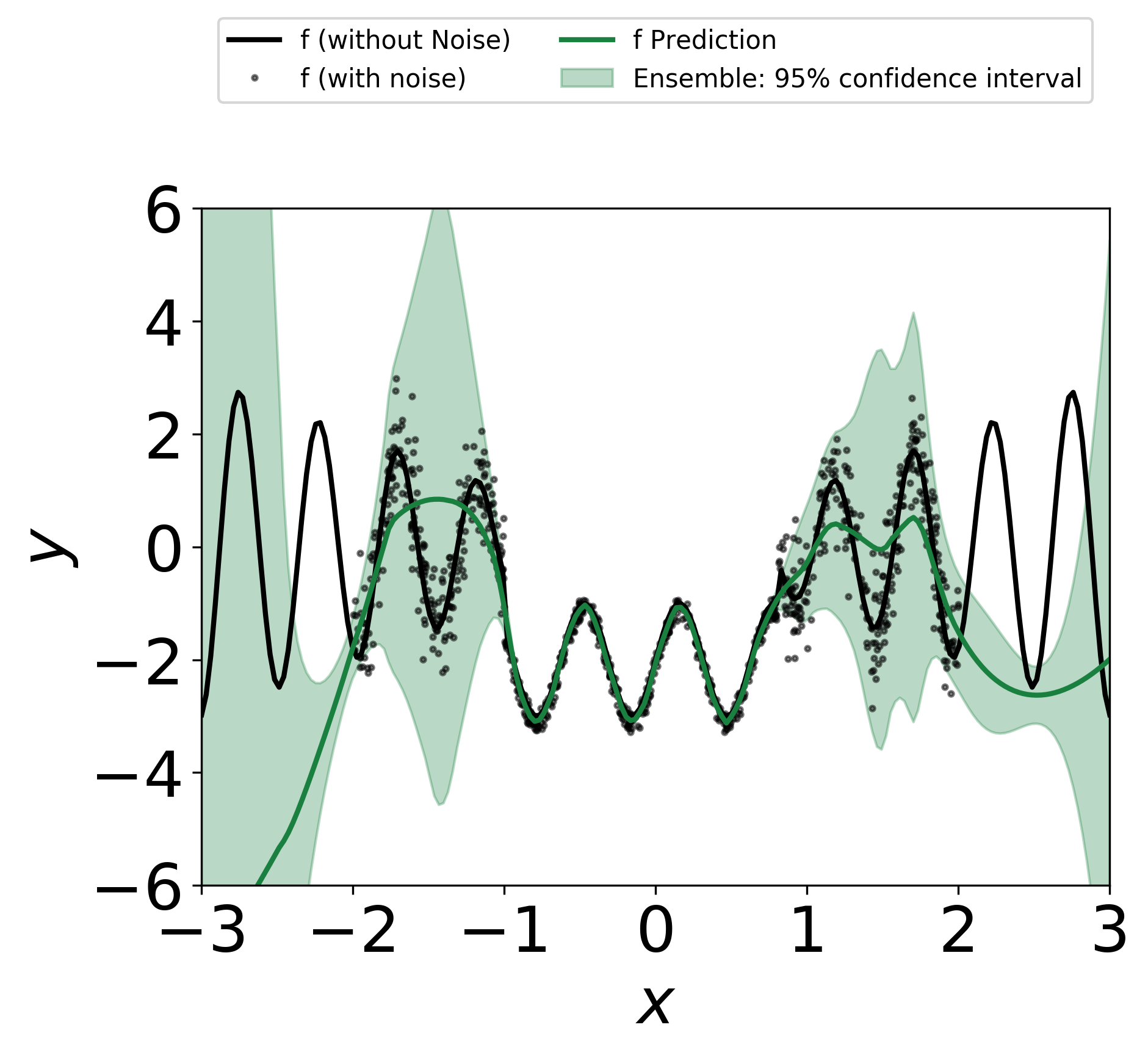}
    }
    \subfloat[MC-Dr, RMSE: 0.48]
    {
        \label{mcdr.fig}
        \includegraphics[width=1.5in,height=1.5in,keepaspectratio,trim={0 0 0 2cm},clip]{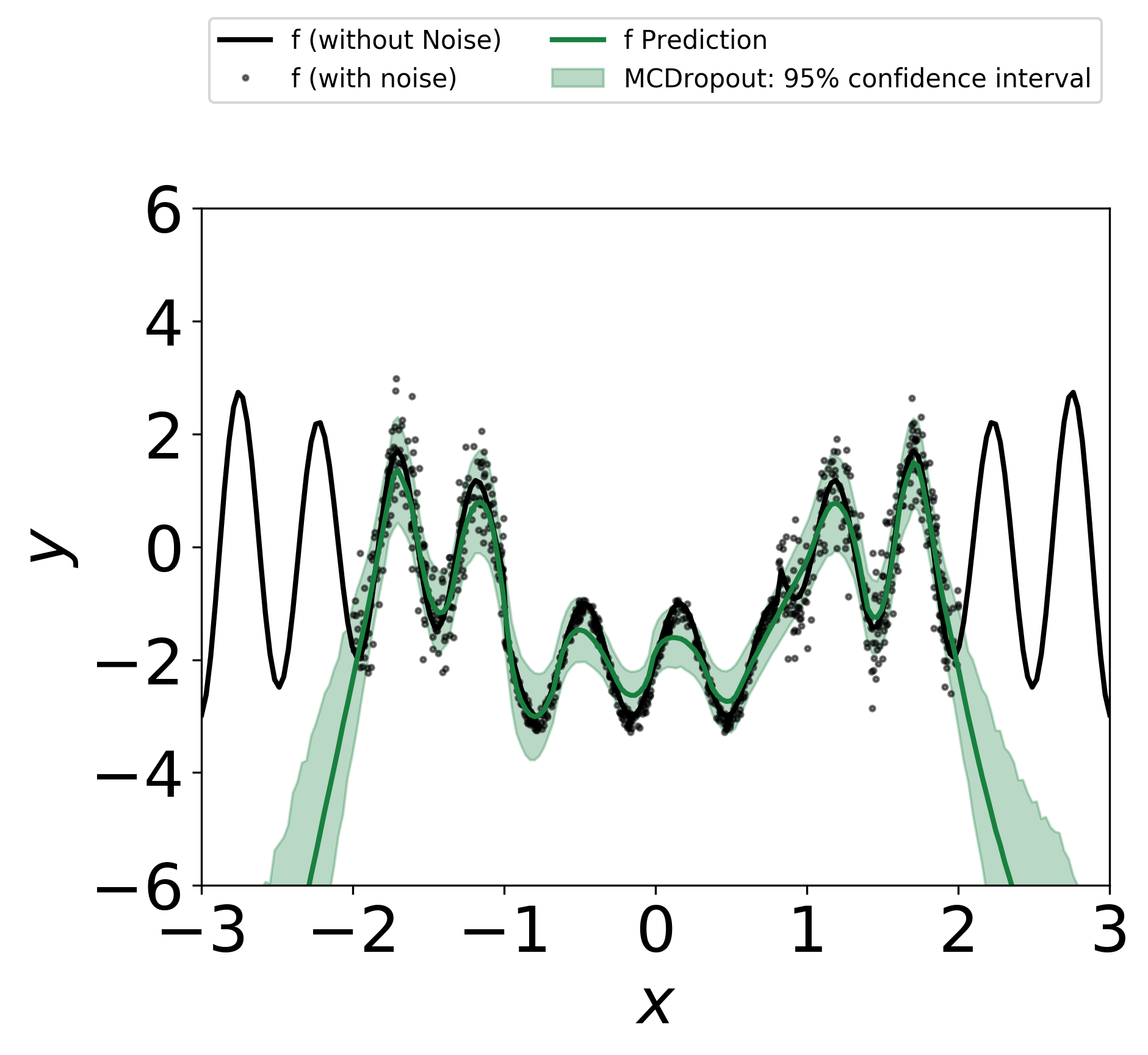}
    }\\
    \subfloat[Split-Train, RMSE: 0.74]
    {
        \label{split.fig}
        \includegraphics[width=1.5in,height=1.5in,keepaspectratio,trim={0 0 0 2cm},clip]{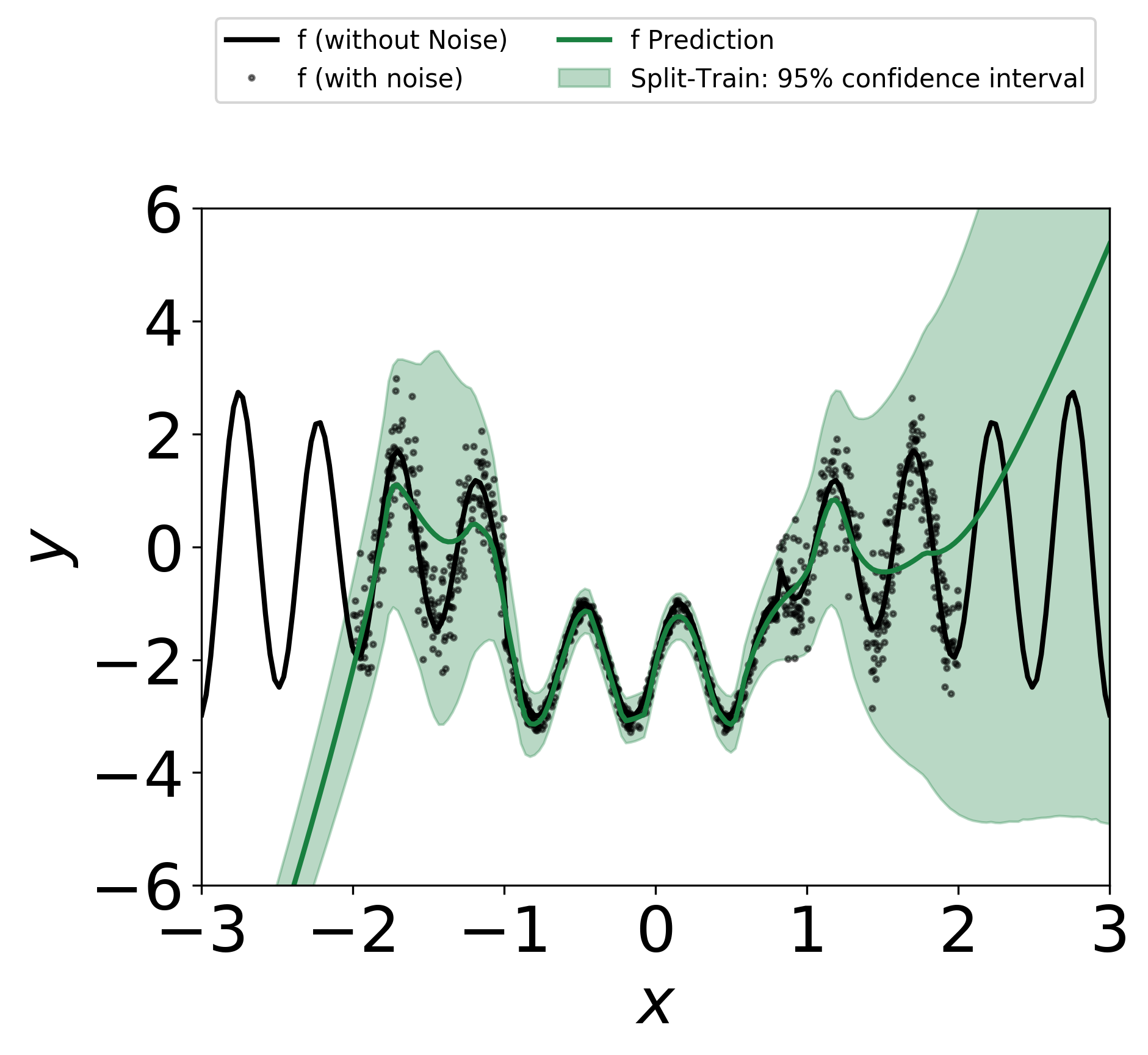}
    }
    \subfloat[DPIN, RMSE: 0.43]
    {
        \label{dvn.fig}
        \includegraphics[width=1.5in,height=1.5in,keepaspectratio,trim={0 0 0 2cm},clip]{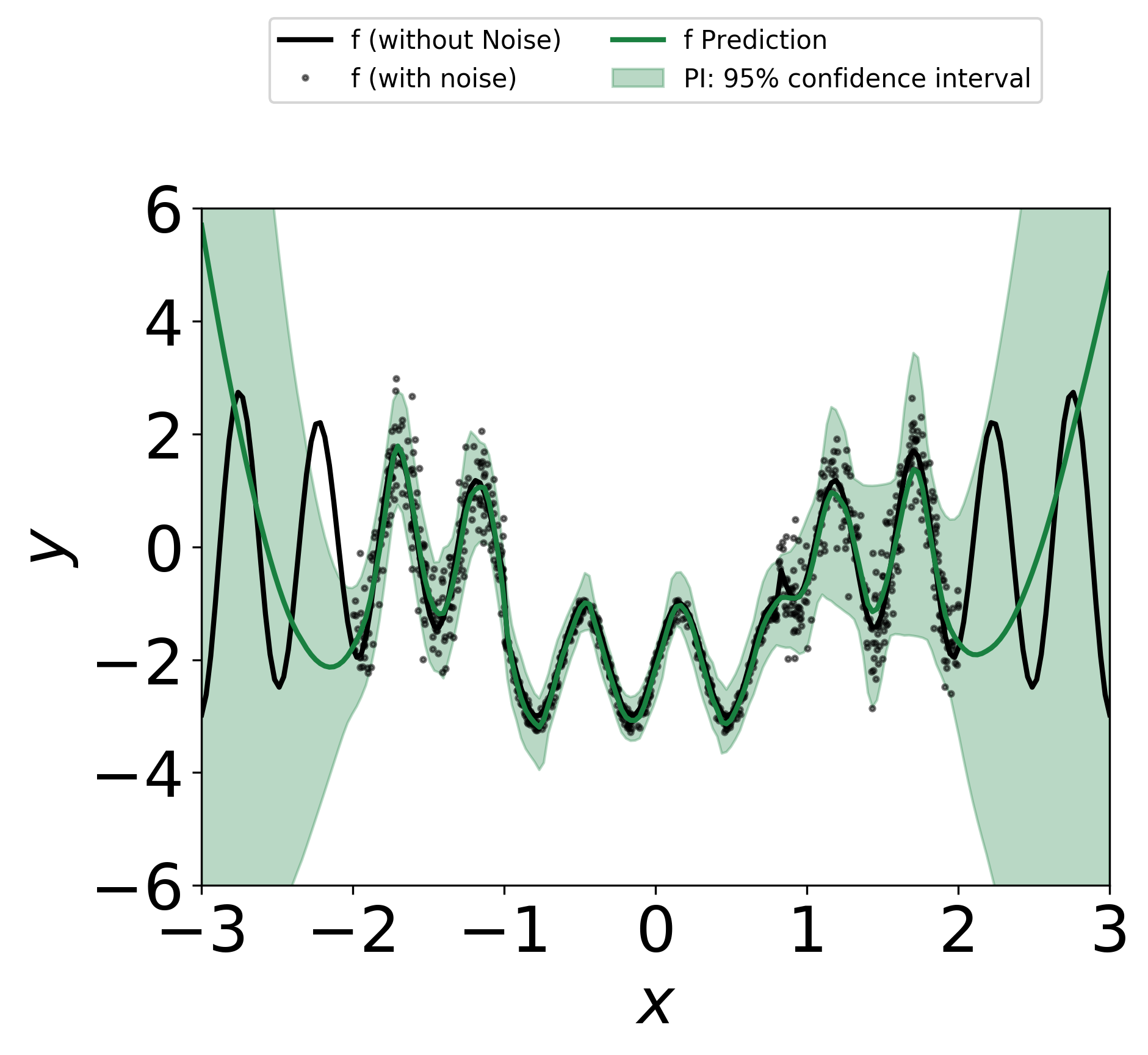}
    }
    \\
    {\vspace{0.3cm} \begin{tikzpicture}[draw=black,scale=0.1]
\draw [line width=1pt] (0,0)(-2,1) -- (2,1);

\end{tikzpicture} F  \quad\quad\quad \begin{tikzpicture}[draw={rgb:red,25;green,124;blue,64},scale=0.1]
\draw [line width=2pt] (0,0)(-2,0.5)-- (4,0.5);

\end{tikzpicture} Target Prediction\\
    \begin{tikzpicture}[draw=black!10,scale=0.1]
\tkzDefPoint(0,0){A}
\foreach \n in {A}
  \node at (\n)[circle,fill,color=black!50,inner sep=1.5pt]{};

\end{tikzpicture} F noise
    \quad\quad \begin{tikzpicture}[draw={rgb:red!30,25;green!30,124;blue!30,64},scale=0.1]
\draw [line width=5pt] (0,0)(-2,0)-- (4,0);

\end{tikzpicture} Prediction Interval}
    \caption{Qualitative prediction accuracy and PI estimation comparison for DPIN, DeepEns, MC-Dr, and Split-Train}
    \label{fig:toy_comp}
    \vspace{-0.2in}
\end{figure}
This experiment demonstrates the ability of DPIN to learn complex functions accurately with varying noise distribution. A total of 1000 data-points are generated using eq. (\ref{eq:fx}) for $\mb{x} \in [-2,2]$ to create a training set for evaluating\footnote{We were not able to produce a Quality-PI solution modelling reasonable approximation for the training data. An extensive hyperparameter search may produce a reasonable solution, but it is out of scope for this paper.} DPIN, DeepEns, Split-Train, and MC-Dr. For this experiment, we measure RMSE and quality of uncertainty estimates.\vspace{-0.1in}
\begin{equation}
    f(x) = 
     \begin{cases}
       \text{$\mathbf{-2 + \sin{(10x+w1)} + w2}$,}\\ \quad\text{\textcolor{darkgray}{$\mb{w1} \sim \mathcal{N}(0.0, 0.0016), \mb{w2} \sim \mathcal{U}(-0.3, 0.1)$}}\\[6pt]
       \text{$\mathbf{x \sin{(12x+w1)} + w2}$,}\\ \text{  }\quad\text{\textcolor{darkgray}{$\mb{w1} \sim \mathcal{U}(-0.4, 0.3), \mb{w2} \sim \mathcal{N}(0.0, 0.25)$}} \\
     \end{cases}
     \label{eq:fx}
\end{equation}  
Figure \ref{fig:toy_comp} represents this comparison between different methods. The DeepEns, in Figure \ref{deepens.fig}, is able to successfully learn the low noise region (lower value of $|x|$), but fails to accurately predict $f(x)$ generated from higher $|x|$ with higher noise. This is confirmed by the higher RMSE of $0.77$ generated with DeepEns predictions on training data. The MC-Dr, in Figure \ref{mcdr.fig}, is able to fairly learn the function values $f(x)$ demonstrating the RMSE of $0.48$ but fails at predicting lower variance for low noise data and higher variance for out of distribution data. The Split-Train, in Figure \ref{split.fig}, showcases similar characteristics as DeepEns, with higher RMSE of $0.74$, owing to similar NLL loss function optimization. DPIN, in Figure \ref{dvn.fig}, demonstrates accurate prediction of the function $f(x)$ with RMSE of $0.43$, lower by 44\% and 41\% compared to DeepEns and Split-Train respectively. It also learns the uncertainty information accurately, predicting lower PI range for lower values of $|x|$, a higher PI range for higher values of $|x|$ within training data, and very high PI for out of distribution data. DPIN does not compromise on prediction accuracy, unlike DeepEns and Split-Train, as well as on variance quality unlike MC-Dr, owing to the decoupled two stage network training, allowing for one to be trained separately while not affecting the other.
                    
\subsection{UCI Dataset Regression}
\begin{table*}[t]
\bgroup
\def\arraystretch{1.3}
\setlength{\tabcolsep}{2pt}
    \centering
    \resizebox{\textwidth}{!}{
    \begin{tabular}{l c|| c| c| c| c| c| c| c| c| c}
    \toprule
    \hline
    \multicolumn{1}{c}{}& \multicolumn{1}{c}{} & \multicolumn{1}{c}{\texttt{Boston}} & \multicolumn{1}{c}{\texttt{Concrete}} & \multicolumn{1}{c}{\texttt{Energy}} & \multicolumn{1}{c}{\texttt{Kin8nm}} & \multicolumn{1}{c}{\texttt{Naval}} & \multicolumn{1}{c}{\texttt{Power}} & \multicolumn{1}{c}{\texttt{Wine}} & 
    \multicolumn{1}{c}{\texttt{Yacht}} & \multicolumn{1}{c}{\texttt{Protein}} \vspace{0.7pt} \\
    \hline
 \multirow{4}{*}{\STAB{\rotatebox[origin=c]{90}{\texttt{RMSE}}}}& MC-Dr & 3.19 $\pm$  0.80 & 5.21 $\pm$  0.61 & 1.29 $\pm$  0.35 & 0.10 $\pm$  0.00 & \textbf{0.00 $\pm$  0.00} & 4.54 $\pm$  0.25 & 0.62 $\pm$  0.05 & 1.16 $\pm$  0.50 & 4.95 $\pm$  0.13\\                                          
& DeepEns & 3.15 $\pm$  0.81 & 5.75 $\pm$  0.55 & 2.36 $\pm$  0.32 & 0.09 $\pm$  0.01 & \textbf{0.00 $\pm$  0.00} & 3.96 $\pm$  0.25 & \textbf{0.61 $\pm$  0.05} & 2.26 $\pm$  0.86 & 4.35 $\pm$  0.05\\ 

& Split-Train & 3.25 $\pm$  0.65 & 6.20 $\pm$  0.96 & 1.38 $\pm$  0.46 & 0.11 $\pm$  0.01 & 0.01 $\pm$  0.00 & 4.43 $\pm$  0.44 & 0.71 $\pm$  0.09 & 1.12 $\pm$  0.60 & 5.75 $\pm$  0.85\\ 

& Quality-PI & 3.73 $\pm$  0.78 & 5.79 $\pm$  0.57 & 2.68 $\pm$  0.25 & 0.09 $\pm$  0.00 & 0.01 $\pm$  0.00 & 4.23 $\pm$  0.24 & 0.72 $\pm$  0.05 & 3.00 $\pm$  1.40 & 5.24 $\pm$  0.07\\ 

& DPIN & \textbf{2.99 $\pm$  0.53} & \textbf{4.49 $\pm$  0.63} & \textbf{0.44 $\pm$  0.06} &\textbf{ 0.07 $\pm$  0.00} & \textbf{0.00 $\pm$  0.00} & \textbf{3.94 $\pm$  0.24} & \textbf{0.61 $\pm$  0.05} & \textbf{0.67 $\pm$  0.27} & \textbf{4.23 $\pm$  0.05}\\
\hline\hline \multirow{4}{*}{\STAB{\rotatebox[origin=c]{90}{\texttt{LL}}}}& MC-Dr & -2.98 $\pm$  0.73 & -3.47 $\pm$  0.28 & -1.73 $\pm$  0.34 & 0.83 $\pm$  0.06 & 5.14 $\pm$  0.16 & -3.30 $\pm$  0.26 & -6.38 $\pm$  1.34 & -2.11 $\pm$  0.86 & -9.80 $\pm$  0.34\\ 

& DeepEns & -2.71 $\pm$  0.39 & -3.02 $\pm$  0.16 & -1.27 $\pm$  0.18 & \textbf{1.21 $\pm$  0.03} & 3.68 $\pm$  0.00 & \textbf{-2.78 $\pm$  0.08} & -1.14 $\pm$  0.47 & -0.94 $\pm$  0.18 & \textbf{-2.75 $\pm$  0.05}\\ 

& Split-Train & -2.57 $\pm$  0.14 & -3.37 $\pm$  0.07 & -2.26 $\pm$  0.06 & 0.86 $\pm$  0.07 & 3.01 $\pm$  0.13 & -3.27 $\pm$  0.02 & -1.04 $\pm$  0.12 & -2.94 $\pm$  0.03 & -3.16 $\pm$  0.13\\ 

& Quality-PI & -2.63 $\pm$  0.22 & -3.10 $\pm$  0.08 & -2.22 $\pm$  0.11 & 1.01 $\pm$  0.03 & 2.89 $\pm$  0.08 & -2.86 $\pm$  0.08 & -1.65 $\pm$  0.25 & -2.66 $\pm$  0.29 & -3.42 $\pm$  0.54\\ 

& DPIN &\textbf{ -2.50 $\pm$  0.24} & \textbf{-2.85 $\pm$  0.12} & \textbf{-0.75 $\pm$  0.08} & 1.19 $\pm$  0.02 &\textbf{ 5.83 $\pm$  0.18} & -2.79 $\pm$  0.06 & \textbf{-0.91 $\pm$  0.10} &\textbf{ -0.90 $\pm$  0.43} & -2.81 $\pm$  0.02\\
\hline\hline 
 
 \multirow{4}{*}{\STAB{\rotatebox[origin=c]{90}{\texttt{PICP}}}}& MC-Dr & 0.83 $\pm$  0.07 & 0.83 $\pm$  0.03 & 0.94 $\pm$  0.04 & 0.88 $\pm$  0.02 & \textbf{1.00 $\pm$  0.00} & 0.84 $\pm$  0.06 & 0.46 $\pm$  0.04 & 0.85 $\pm$  0.08 & 0.38 $\pm$  0.02\\ 

& DeepEns & 0.84 $\pm$  0.05 & 0.92 $\pm$  0.02 & \textbf{0.97 $\pm$  0.02} & \textbf{0.96 $\pm$  0.01} & \textbf{1.00 $\pm$  0.00} & \textbf{0.96 $\pm$  0.01} & 0.91 $\pm$  0.02 & \textbf{0.99 $\pm$  0.02} &\textbf{ 0.96 $\pm$  0.00}\\ 

& Split-Train & \textbf{0.96 $\pm$  0.03} & \textbf{0.99 $\pm$  0.01 }& \textbf{1.00 $\pm$  0.00} & 0.94 $\pm$  0.02 & \textbf{0.97 $\pm$  0.04} & \textbf{1.00 $\pm$  0.00} & 0.92 $\pm$  0.03 & \textbf{1.00 $\pm$  0.00} & \textbf{0.96 $\pm$  0.01}\\ 

& Quality-PI & \textbf{0.97 $\pm$  0.03} & \textbf{0.97 $\pm$  0.02} & \textbf{1.00 $\pm$  0.00} & \textbf{0.97 $\pm$  0.01} & \textbf{0.99 $\pm$  0.01} & \textbf{0.96 $\pm$  0.01} &\textbf{ 0.94 $\pm$  0.02} & \textbf{1.00 $\pm$  0.00} & \textbf{0.95 $\pm$  0.00}\\ 

& DPIN & 0.90 $\pm$  0.04 & \textbf{0.95 $\pm$  0.02} & \textbf{0.99 $\pm$  0.01} & \textbf{0.99 $\pm$  0.00 }& \textbf{1.00 $\pm$  0.00} & \textbf{0.97 $\pm$  0.01} & 0.92 $\pm$  0.03 &\textbf{ 0.96 $\pm$  0.04} & \textbf{0.95 $\pm$  0.00}\\

\hline\hline
 \multirow{4}{*}{\STAB{\rotatebox[origin=c]{90}{\texttt{MPIW}}}}& MC-Dr & 0.86 $\pm$  0.47 & 1.02 $\pm$  0.36 & 0.56 $\pm$  0.20 & 1.10 $\pm$  0.28 & 0.63 $\pm$  0.18 & 0.85 $\pm$  0.34 & 0.92 $\pm$  0.34 & 0.37 $\pm$  0.34 & 0.96 $\pm$  0.31\\ 

& DeepEns & 0.86 $\pm$  0.08 & 1.04 $\pm$  0.06 & 0.57 $\pm$  0.07 & \textbf{1.24 $\pm$  0.03} & \textbf{0.23 $\pm$  0.02} & \textbf{0.88 $\pm$  0.01} & 2.52 $\pm$  0.10 & 0.42 $\pm$  0.13 & 2.69 $\pm$  0.04\\ 

& Split-Train & 1.70 $\pm$  0.18 & 2.18 $\pm$  0.02 & 1.36 $\pm$  0.01 & 1.55 $\pm$  0.07 & 3.33 $\pm$  0.35 & 2.16 $\pm$  0.00 & 2.88 $\pm$  0.24 & 1.92 $\pm$  0.03 & 3.93 $\pm$  0.29\\ 

& Quality-PI & \textbf{1.58 $\pm$  0.17} & 1.39 $\pm$  0.06 & 1.10 $\pm$  0.13 & 1.51 $\pm$  0.02 & 3.22 $\pm$  0.16 & 0.89 $\pm$  0.01 & \textbf{2.06 $\pm$  0.07} & 1.38 $\pm$  0.30 & \textbf{2.40 $\pm$  0.03}\\ 

& DPIN & 0.95 $\pm$  0.04 &\textbf{ 1.00 $\pm$  0.03} & \textbf{0.27 $\pm$  0.02} & 1.35 $\pm$  0.02 & 0.30 $\pm$  0.07 & 0.94 $\pm$  0.01 & 2.65 $\pm$  0.08 & \textbf{0.18 $\pm$  0.02} & 2.60 $\pm$  0.03\\
\toprule
\end{tabular}
} 
\caption{Comparison of RMSE, LL, PICP and MPIW between different algorithms across multiple datasets. The results are presented in mean $\pm$ one standard error with best results are emphasized with bold. RMSE is expected to be as low as possible, LL should be as high as possible. PICP is expected to be $\geq 95\%$, and MPIW should be as low as possible. Best results are highlighted for all the metrics. The highlighted MPIW is lowest when compared between methods generating $PICP \geq 95\%$ or with highest PICP value (in case none of the method is able to achieve $PICP \geq 95\%$).}
    \label{tab:mcdr_split_dvn}
\egroup
\vspace{-0.2in}
\end{table*}
We now test DPIN across 9 publicly available datasets from UCI benchmark. We follow the experimental setup of \cite{prob-backprop} to compare the performance of different algorithms. The experiments for all the datasets are performed 20 times, and average results are presented subsequently. The prediction accuracy can be evaluated by comparing the RMSE of the test set, but the evaluation metrics for estimating the reliability of the predictive uncertainty varies across the literature. We have demonstrated the evaluation of prediction accuracy using RMSE and quality of uncertainty using 1. LL, 2. $95\%$ PICP, and 3. Mean Prediction Interval Width (MPIW). The $95\%$ PICP quantifies the percentage of observations falling within $95\%$ of prediction interval (PI) and MPIW represents the average width of the PI. The best model is the one with least RMSE, highest LL, PICP as close to $95\%$ as possible with minimum MPIW. The RMSE, LL, PICP and MPIW comparison between MC-Dr, Split-train\footnote{Reproduced using \tiny{\url{https://github.com/SkafteNicki/john}}}, Quality-PI\footnote{Reproduced using \tiny{\url{https://github.com/TeaPearce/Deep_Learning_Prediction_Intervals.git}}} and DPIN are shown in Table \ref{tab:mcdr_split_dvn}. 

The best RMSE (lower is better) is \textbf{emphasized}, representing the best possible scenario for the network predictions. DPIN is able to achieve better or similar accuracy over all the datasets, except one. It is also able to reduce RMSE of the predictions by approximately $16\%$ compared to MC-Dr, $19\%$ compared to DeepEns, $31\%$ compared to Split-Train and $37\%$ compared to Quality-PI on average.

The best LL (higher is better) is \textbf{emphasized}, representing the quality of predictive uncertainty, achieving high variance for high error estimates, and low variance for low error estimates. Similar to Quality-PI, we computed the standard deviation for LL with the equivalent of the Gaussian distribution for PI ($(\lambda_U+\lambda_L)/3.92$). DPIN is able to attain higher LL for 6 out of 9 UCI datasets. DPIN showcases LL improvements of approximately $35\%$ compared to MC-Dr, $10\%$ compared to DeepEns, $19\%$ compared to Split-Train, and $24\%$ compared to Quality-PI on average.

 Each instance satisfying the condition of $PICP \geq 95\%$ is \textbf{emphasized} in the table. When none of the methods generate $PICP \geq 95\%$, the algorithm with the largest PICP is highlighted to represent the best-case scenario. DPIN is able to achieve required PICP for most of the datasets. Where PICP does narrowly loose out (\texttt{Boston} and \texttt{Wine}), the mean PI range with one standard deviation demonstrates a coverage of $94\%$ and $95\%$, respectively.

The comparison for MPIW depends on the PICP of the target labels. MPIW between different algorithms is compared only when the mean PICP of the predicted uncertainty is greater than or at least equal to $95\%$. Compared to other state-of-the-art methods, DPIN is able to achieve the least MPIW in 3 datasets (\texttt{Concrete}, \texttt{Energy}, and \texttt{Yacht}), second least in 3 datasets (\texttt{kin8nm}, \texttt{Naval}, and \texttt{Protein}) and third least in 1 dataset (\texttt{Power}) out of 7 datasets with desired PICP coverage.

\begin{table*}[t]
\bgroup
\def\arraystretch{1.3}
\setlength{\tabcolsep}{2pt}
    \centering
    \resizebox{\textwidth}{!}{
    \begin{tabular}{l|| c| c| c| c| c| c| c| c|c}
    \toprule
    \hline
 \multicolumn{1}{c}{} & \multicolumn{1}{c}{\texttt{Boston}} & \multicolumn{1}{c}{\texttt{Concrete}} & \multicolumn{1}{c}{\texttt{Energy}} & \multicolumn{1}{c}{\texttt{Kin8nm}} & \multicolumn{1}{c}{\texttt{Naval}} & \multicolumn{1}{c}{\texttt{Power}} & \multicolumn{1}{c}{\texttt{Wine}} & 
    \multicolumn{1}{c}{\texttt{Yacht}} & \multicolumn{1}{c}{\texttt{Protein}} \vspace{0.7pt} \\
    \hline
MC-Dr & 3.31 $\pm$  0.48 & 6.28 $\pm$  0.33 & 2.02 $\pm$  0.17 & 0.11 $\pm$  0.00 & 0.01 $\pm$  0.00 & 4.20 $\pm$  0.06 & \textbf{0.63 $\pm$  0.02 }& 1.96 $\pm$  0.31 & 4.63 $\pm$  0.03\\                                          
DeepEns & 3.44 $\pm$  0.66 & 6.15 $\pm$  0.45 & 2.36 $\pm$  0.15 & 0.11 $\pm$  0.01 & 0.01 $\pm$  0.00 & 4.05 $\pm$  0.04 & 0.64 $\pm$  0.02 & 1.07 $\pm$  0.29 & 4.67 $\pm$  0.04\\ 

Split-Train & 4.33 $\pm$  1.04 & 9.05 $\pm$  1.95 & 1.40 $\pm$  0.24 & 0.18 $\pm$  0.03 & 0.01 $\pm$  0.01 & 4.66 $\pm$  0.53 & 0.88 $\pm$  0.20 & 1.60 $\pm$  0.41 & 6.52 $\pm$  1.69\\ 

DPIN & \textbf{3.23 $\pm$  0.41} & \textbf{5.50 $\pm$  0.17} & \textbf{0.64 $\pm$  0.07} & \textbf{0.08 $\pm$  0.00} & \textbf{0.00 $\pm$  0.00} & \textbf{4.03 $\pm$  0.05} & 0.66 $\pm$  0.03 & \textbf{0.82 $\pm$  0.22} & \textbf{4.52 $\pm$  0.03}\\
\hline
\end{tabular}
}
\caption{Average RMSE test error and standard error in active learning}
    \label{tab:active}
\egroup
\vspace{-0.2in}
\end{table*}
\subsection{Active Learning}
\label{sec:active}
We also evaluate the quality of uncertainty estimates using active learning. The network predictions for output significantly depend on the quality of variance in active learning \cite{activelearn}. In these experiments, we use the same network architecture and datasets as in the UCI benchmarking. We split each dataset into $30\%$ train, $50\%$ pool, and $20\%$ test data. For each active learning iteration, the network is optimized using training data, and the prediction accuracy of the test data is evaluated. After each active learning iteration, $n = 10$ samples with the highest predicted uncertainty (corresponding to highest entropy \cite{active_var}) from the pool data are moved to the training dataset to perform the next iteration. We perform a total of $10$ active learning iterations producing  $10$ evaluations on the test set. The entire process from data splitting to active learning iterations are repeated $10$ times to acquire more robust estimates.
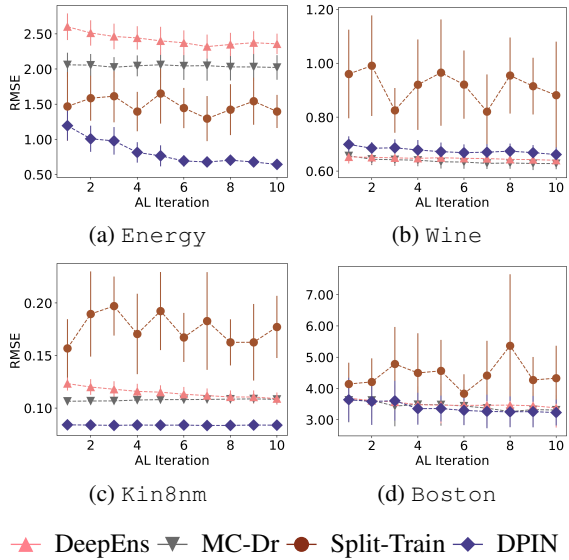
\begin{figure}[t]
    \centering
    \subfloat[\texttt{Energy}]
    {
        \label{fig:active_energy}
        \includegraphics[width=1.5in,keepaspectratio]{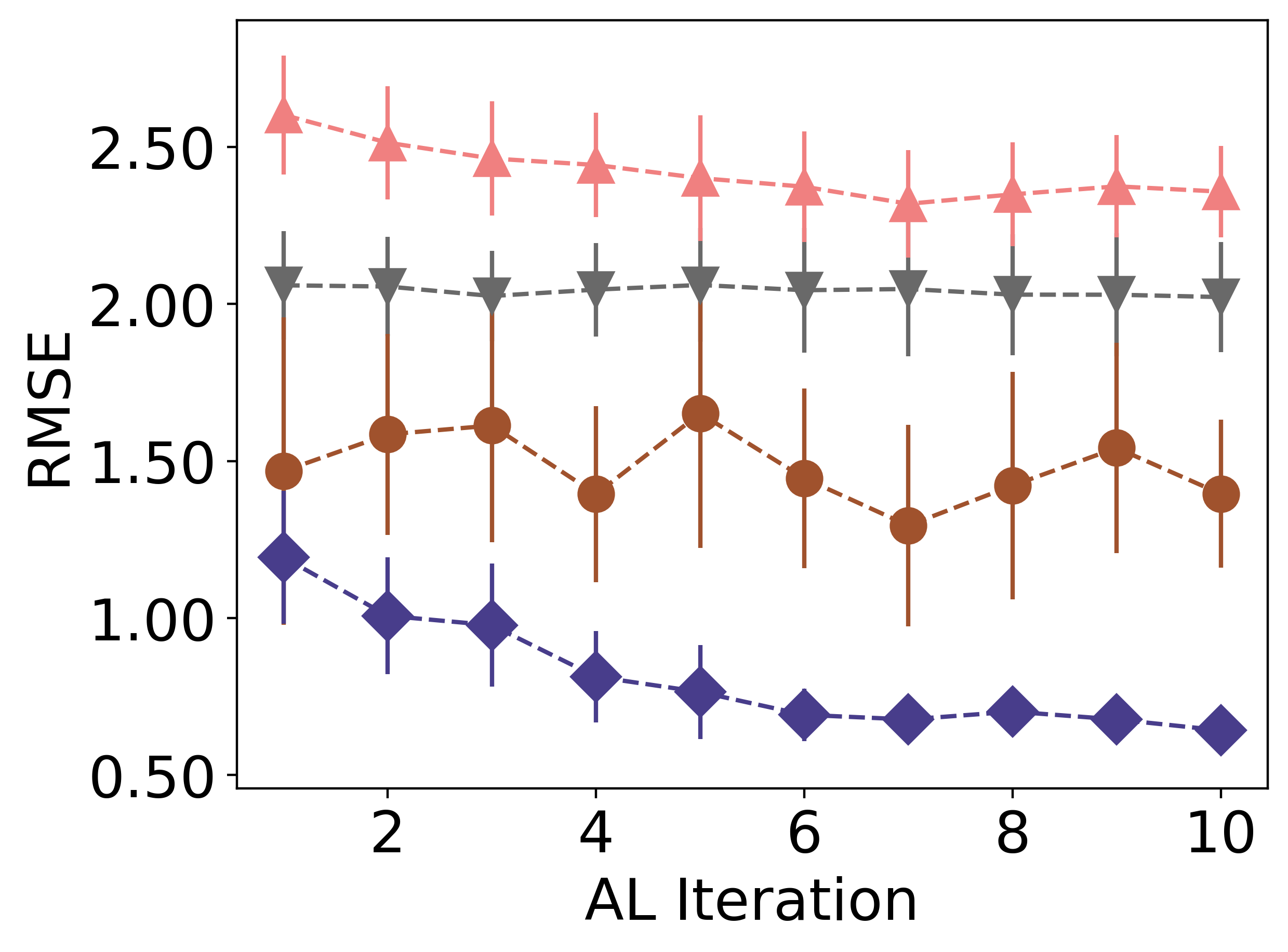}
    }
    \subfloat[\texttt{Wine}]
    {    
        \label{fig:active_wine  }
        \includegraphics[width=1.39in,keepaspectratio,trim={1cm 0 0 0},clip]{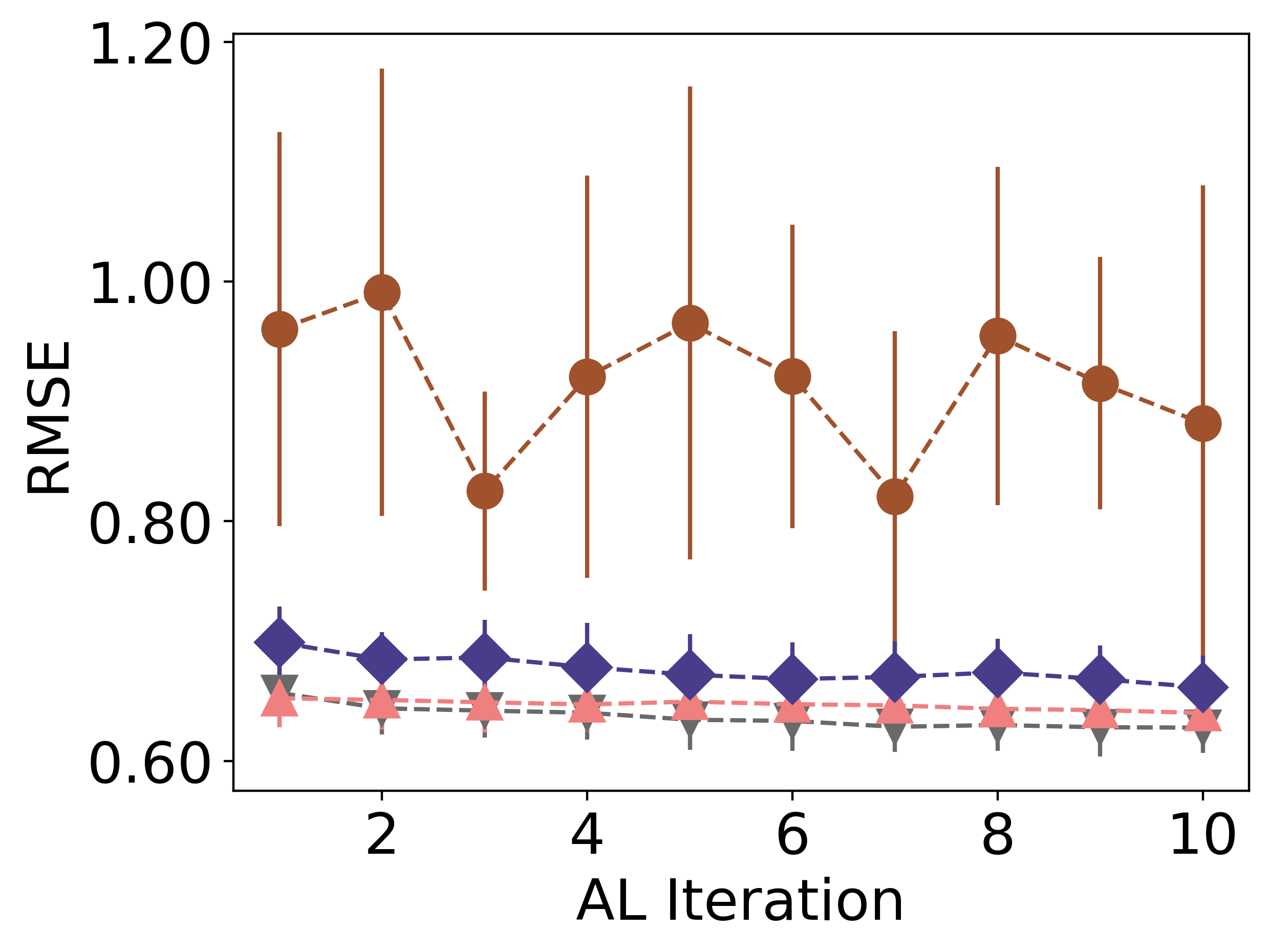}
    }
    \\ \vspace{-0.1in}
    \subfloat[\texttt{Kin8nm}]
    {
        \label{fig:active_kin8nm}
        \includegraphics[width=1.5in,keepaspectratio]{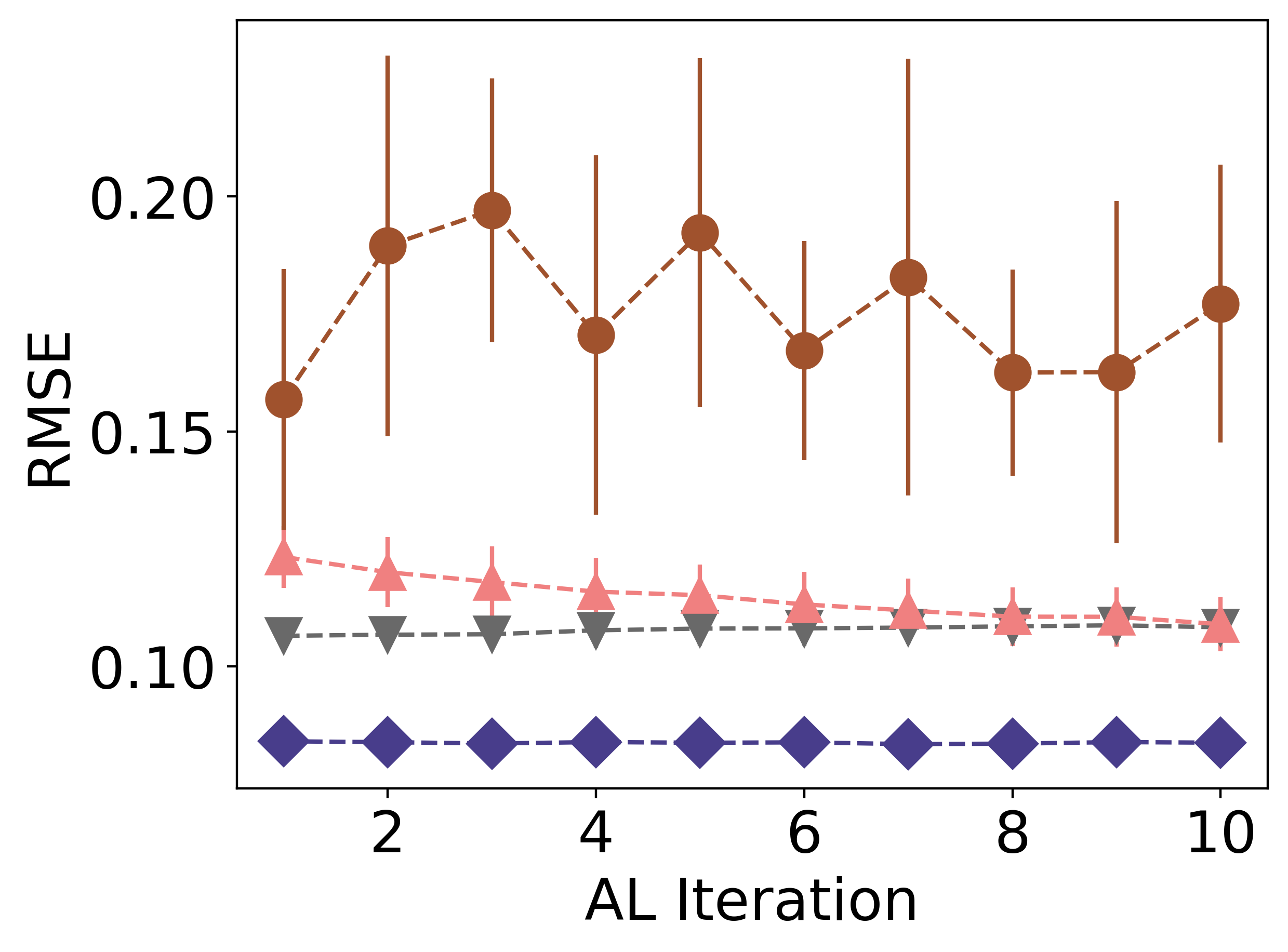}
    }
    \subfloat[\texttt{Boston}]
    {
        \label{fig:active_boston}
        \includegraphics[width=1.39in,keepaspectratio,trim={1cm 0 0 0},clip]{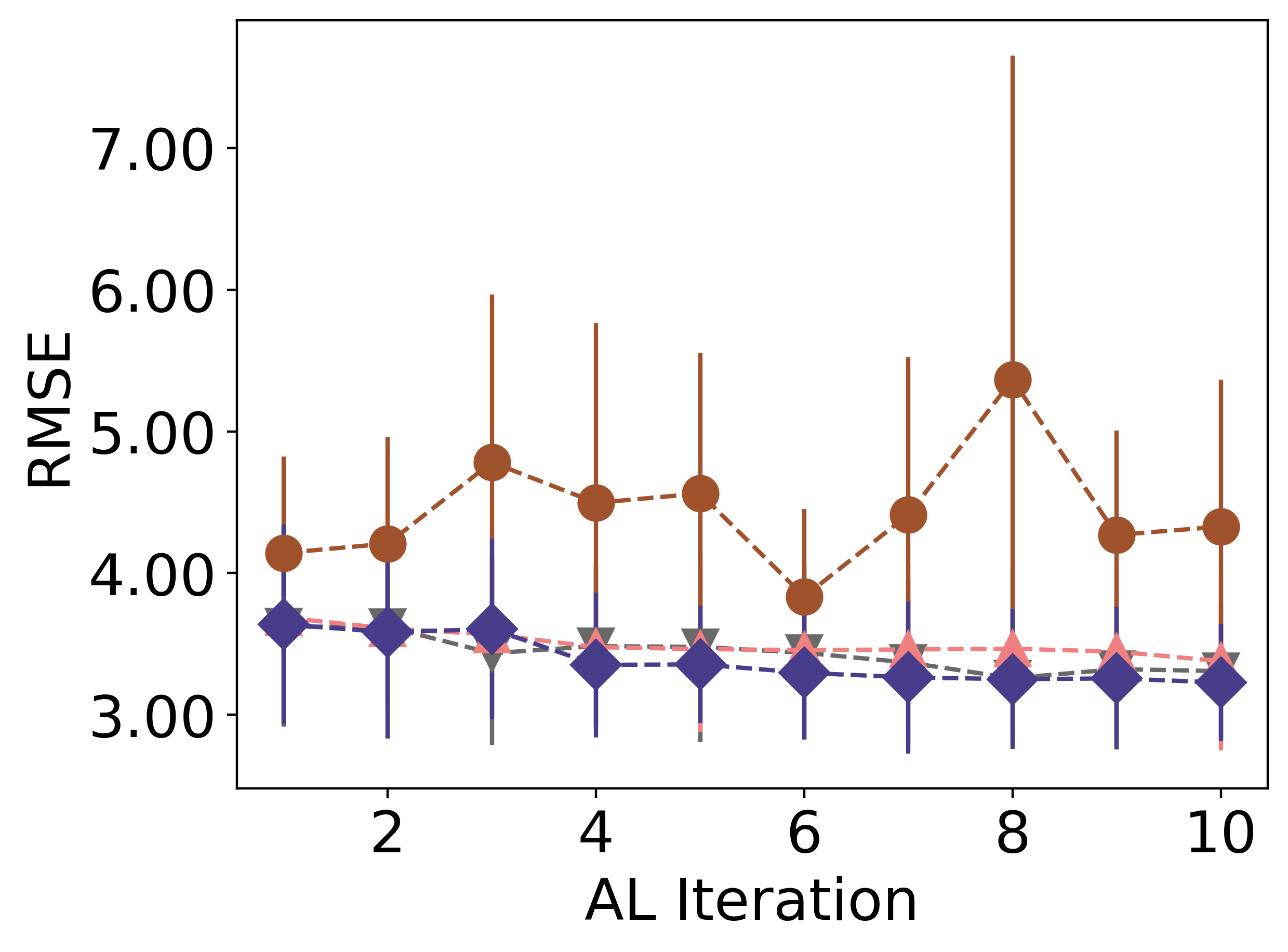}
    }
    \\
    {\vspace{0.3cm} \begin{tikzpicture}[scale=0.1]
\draw [line width=0.5pt,coralred] (-1,1)-- (3,1);

\fill [coralred] (0,0) -- (2,0) -- (1,2) -- (0,0);

\end{tikzpicture} DeepEns \begin{tikzpicture}[scale=0.1]
\draw [line width=0.5pt,dimgray] (-2,1)-- (2,1);

\fill [dimgray] (0,0) -- (-1,2) -- (1,2) -- (0,0);

\end{tikzpicture} MC-Dr \begin{tikzpicture}[scale=0.1]
\draw[line width=0.5pt,sienna] (-2,0)-- (2,0);

\fill[sienna](0,0) circle (1cm);

\end{tikzpicture} Split-Train \begin{tikzpicture}[scale=0.1]
\draw [line width=0.5pt,darkslateblue] (-2,1)-- (2,1);
\fill [darkslateblue] (0,0) -- (-1,1) -- (0,2) -- (1,1) -- (0,0) --(1,0);

\end{tikzpicture} DPIN 
    }
    \caption{Evolution of Average RMSE test error and standard error in the active learning experiment over 10 learning iterations.}
    \label{fig:active}
    \vspace{-0.2in}
\end{figure}
Table \ref{tab:active} summarises the average test RMSE error at the end of the data collection process, i.e., after $10^{th}$ active learning iteration, for MC-Dr, DeepEns, Split-Train, and DPIN methods. The Quality-PI algorithm does not produce a valid PI range for multiple active learning iterations. Further efforts involved are out of scope for this paper. The active learning results for DPIN showcase significant improvement in the RMSE error for all the datasets except \texttt{Wine} over other methods. DPIN demonstrates RMSE reduction by approximately $21\%$ compared to MC-Dr, $17\%$ compared to DeepEns, and $36\%$ compared to Split-Train for network predictions, indicating the effective learning of predictive uncertainty.

Figure \ref{fig:active} demonstrates the evolution of average test RMSE error with standard error for different algorithms (DeepEns, MC-Dr, Split-Train, and DPIN) across datasets: \texttt{Energy}, \texttt{Wine}, \texttt{Kin8nm} and \texttt{Boston}. For datasets \texttt{Energy} and \texttt{Kin8nm}, the proposed decoupled approach is able to achieve significantly better accuracy early on; by upto 54\% and 50\% respectively, at iteration 1. We continue to improve as more data points are added from the pool for an RMSE reduction of upto 73\% on \texttt{Energy} and 55\% on \texttt{Kin8nm} datasets, at iteration 10. For other datasets like \texttt{Wine}, and \texttt{Boston}, our proposed method is on par with the other state-of-the-art approaches, achieving similar accuracy and maintaining the learning trend. Evolution of average RMSE for remaining datasets and average LL for all the datasets is included in the supplementary material.

\section{Conclusion}
In this paper, we demonstrate a two-stage training approach, with decoupled networks modeling target predictions and corresponding PIs. Our decoupled target prediction network achieves state-of-the-art prediction accuracy with reliable uncertainty estimation. We derive a custom distribution-free loss function for characterizing the aleatoric uncertainty of the estimates. It optimizes the PI range around target predictions, with the desired PICP covering the proportion of the target labels in PI estimations. The proposed solution is evaluated and compared against current uncertainty estimation approaches on a synthetic dataset and UCI benchmark. DPIN is able to reduce the RMSE by 18\% on average while improving the LL by 8\% across multiple datasets. The proposed method is able to maintain $95\%$ coverage probability on 7 out of 9 UCI datasets with comparable PI widths for UCI regression related tasks. Furthermore, the quality of PI is examined by comparing the performance of the methods on active learning, where DPIN outperforms other algorithms with an average RMSE reduction of 14\%, demonstrating reliable PI modelling.


\bibliography{example_paper}
\bibliographystyle{icml2020}

\end{document}